\documentclass[review]{elsarticle}

\usepackage{lineno,hyperref}
\usepackage{amsmath,amssymb}
\usepackage{booktabs}
\usepackage{subfig}
\modulolinenumbers[5]

\journal{Journal of \LaTeX\ Templates}

\begin{document}

\begin{frontmatter}

\title{A survey on trajectory clustering analysis}

\author[mymainaddress]{Jiang Bian}
\ead{bianjiang22@gmail.com}

\author[mysecondaryaddress]{Dayong Tian\corref{mycorrespondingauthor}}
\cortext[mycorrespondingauthor]{Corresponding author}
\ead{dayongt@gmail.com}

\author[mytertiaryaddress]{Yuanyan Tang}
\ead{yytang@umac.mo}

\author[myfourthaddress]{Dacheng Tao}
\ead{dacheng.tao@sydney.edu.au}

\address[mymainaddress]{Centre for Artificial Intelligence, FEIT, University of Technology Sydney, Australia}
\address[mysecondaryaddress]{School of Electronics and Information, Northwestern Polytechnical University, China}
\address[mytertiaryaddress]{Faculty of Science and Technology, University of Macau, China}
\address[myfourthaddress]{UBTECH Sydney AI Centre, FEIT, University of Sydney, Australia}

\begin{abstract}
This paper comprehensively surveys the development of trajectory clustering. Considering the critical role of trajectory data mining in modern intelligent systems for surveillance security, abnormal behavior detection, crowd behavior analysis, and traffic control, trajectory clustering has attracted growing attention. Existing trajectory clustering methods can be grouped into three categories: unsupervised, supervised and semi-supervised algorithms. In spite of achieving a certain level of development, trajectory clustering is limited in its success by complex conditions such as application scenarios and data dimensions. This paper provides a holistic understanding and deep insight into trajectory clustering, and presents a comprehensive analysis of representative methods and promising future directions.
\end{abstract}


\end{frontmatter}

\section{Introduction}
\begin{figure} [!h]
\centering
\includegraphics[width=2.8in,height=0.5in]{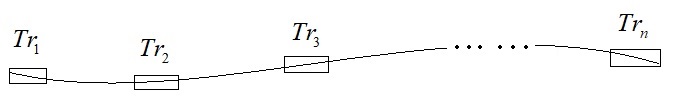}
\caption{Trajectory generated by GPS tracking devices}
\label{fig1}
\end{figure}

\begin{figure}[!h]
\centering
\includegraphics[width=2.6in,height=1.5in]{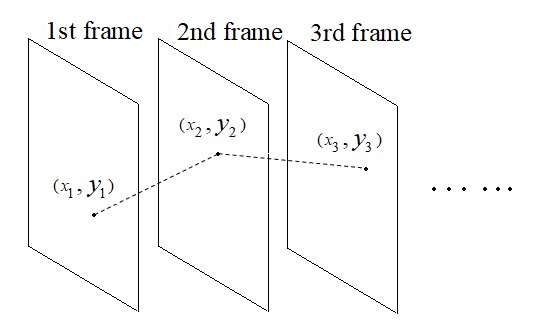}
\caption{Trajectory generated from camera device}
\label{fig2}
\end{figure}

\indent With the development of tracking and surveillance devices, tremendous numbers of object trajectory data were collected, which makes extracting useful information imperative and challenging. Trajectory clustering is an efficient method to analyze trajectory data and it has been applied in pattern recognition, data analysis and machine learning, etc. Furthermore, trajectory clustering aims to gain space, time or even potential information inside trajectory data, so it is ubiquitous in some application fields such as object motion prediction \cite{chen2010searching}, traffic monitoring \cite{atev2006learning}\cite{li2007traffic}\cite{gurung2014traffic}, activity understanding \cite{bashir2007object}\cite{wang2016globally}\cite{yao2017learning}, abnormal detection \cite{wang2014detection}\cite{yuan2017statistical}\cite{zhao2018gestalt}\cite{chaker2017social}, 3 dimensional reconstruction\cite{kumar2017spatio}, weather forecasting \cite{ferreira2013vector} and geography \cite{mo2015application}.\\
\indent Trajectory data are recorded in different formats according to device types, object movements and even purposes. For instance, GPS tracking devices generate a trajectory by tracking object movement as $Trajectory=(Tr_1,Tr_2,\cdots,Tr_n)$, which is a consecutive sequence of points in geographical space, and $Tr_i$ denotes a combination of coordinates and time stamp like $Tr_i=(x_i,y_i,t_i)$, as shown in Fig.\ref{fig1}. In some specific circumstances, other properties relevant to object movement are added, such as velocity, direction or acceleration. As shown in Fig.\ref{fig2}, for image data, a sequence of pixels in consecutive frames forms up a trajectory, which is similar to optical flow \cite{cai2016framework}\cite{wang2016robust}.\\
\indent In order to measure similarities among different types of trajectory data, data representation, feature extraction and distance metric selection are critical preliminary works of trajectory clustering. For example, trajectories can be represented as a vector and downsampled to an unified length, so Euclidean distance is used \cite{nanni2006time}. Trajectories also can be treated as samples of a probablistic distribution. Hence, Bhattacharyya Distance \cite{li2006coarse} is used to measure the distance between two distributions.\\
\indent According to the availability of labeled data, trajectory clustering methods are divided in three categories: unsupervised, supervised, semi-supervised. Unsupervised models aim at clustering data without human experts supervision or labeled data. An inference function has been drawn by analyzing unlabeled data sets \cite{ester1996density}\cite{ferreira2013vector}\cite{wang2006learning}\cite{xiang2008spectral}. Supervised models are learned prior to trajectory clustering. Generally, labeled data are used to learn a function mapping data to their labels, i.e. clusters. The clusters of unlabeled data are predicted by this function, then \cite{gao2007efficient}\cite{yuan2017statistical}\cite{wu2016harnessing}\cite{cho2014classifying}. Labeling data need a heavy burden of manual works by human experts. It is unfeasible for large data sets. Semi-supervised compromises the previous two types of models. It is trained by labeled data and tuned by unlabeled data \cite{gurung2014traffic}\cite{wang2014detection}\cite{yuan2017statistical}.\\
\indent The rest of this paper is organized as follows. Preliminary works are introduced in Sect.2, and the models based on unsupervised algorithms are described in Sect.3. A description of the models under supervised algorithms are presented in Sect.4. Sect.5 discusses some models based on semi-supervised algorithms. Finally, promising future directions are given in Sect.6, and conclusions are made in Sect.7.
\section{Preliminaries}
\subsection{Trajectory Clustering Preparation}
\begin{figure}[!h]
\centering
\includegraphics[width=2.4in,height=1.4in]{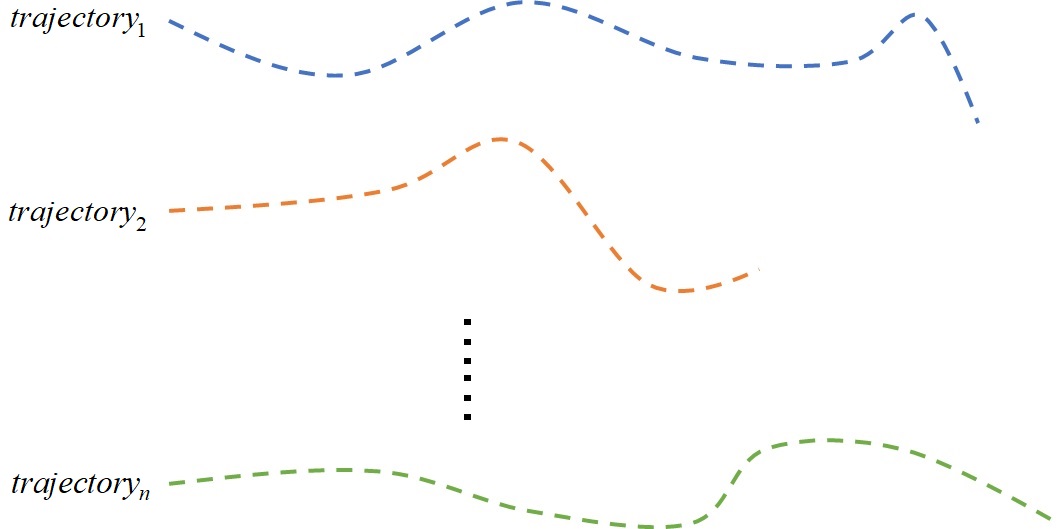}
\caption{Trajectory data with varying lengths}
\label{fig3}
\end{figure}

\indent In some clustering models \cite{zhang2009learning}\cite{sillito2008semi}\cite{naftel2006motion}\cite{hu2013incremental}, trajectory data are required to be set as a unified length so that they could be measured. However, as shown in Fig.\ref{fig3}, for two arbitrary trajectories, their lengths maybe largely different from each other. Therefore, representing trajectories in an unified length with little loss of information is a major preliminary work of these models. This procedure is called clustering preparation.\\
\subsubsection{Trajectory Transformation Algorithms}
\indent For some methods, original data are represented in other space with a same length. For instance, trajectory data are projected into a subspace \cite{hu2015exploiting}. Linear transformation algorithm aims at representing trajectory as a combination of basis trajectories \cite{akhter2011trajectory}. Curve fitting is another method to approximate trajectories by a parameterized quadratic curve \cite{zhang2009learning}. In order to distinguish similar curves, the direction of the last trajectory point is chosen as an additional parameter. In \cite{sillito2008semi}, trajectory data are approximated by a uniform cubic B-spline curve, so that a representation capable of encoding both the shape and the spatiotemporal profile of trajectory data is obtained. In addition, the lengths of trajectories are added to distinguish the trajectories with similar shapes. According to the fact that trajectory data contain a lot kinds of positional information, such as coordinates, speed and directions, vector fields are employed to represent trajectory data \cite{ferreira2013vector}. Vector fields give trajectory a smooth streamline and induce a notion of similarity of trajectories. Principle Component Analysis (PCA) is a statistical procedure to compute a set of linearly uncorrelated variables called principle components by orthogonal transformation. To avoid partially extracted information, a number of organized segmentations substitute for the corresponding trajectory in \cite{bashir2007object} and \cite{bashir2007real}. The time ordering data are transformed and represented in frequency domain by Discrete Fourier Transformation (DFT), so a trajectory can be represented as a fixed length vector comprised of Fourier coefficients in \cite{naftel2006motion} and \cite{hu2013incremental}. In \cite{jiang2012trajectory}, the interaction of trajectories are encoded and set as elements of codebook, so camera motion is ignored and the model's robustness is improved.\\
\subsubsection{Re-sampling Methods}
\indent Re-sampling methods choose trajectory points by sampling rule to unify trajectory lengths. Trajectory data are segmented as sub-trajectories, and all of them are re-sampled to a fixed length so that sub-trajectories are aligned as matrix \cite{bashir2007real}. In a complex scene such as hand writing data set, Equidistant sampling fixes the problem that two same characters are recorded in different temporal sequence because of different writing speeds \cite{schenk2006novel}. Since re-sampled trajectory points are discontinuous, it is critical that normalization should be involved after re-sampling \cite{liou2011trajectory}. It has been widely acknowledged that re-sampling method causes information loss \cite{piotto2009syntactic}. Therefore, sparsity regularization is used in \cite{wang2009evaluation}\cite{ochs2014segmentation}\cite{elhamifar2009sparse}\cite{cappe2007overview}.\\
\subsubsection{Trajectory Substitute}
\indent Sub-trajectories hold partial and hidden information of original trajectory data \cite{hu2013incremental}\cite{lee2007trajectory}, so they are put together and describe trajectory more flexible. For instance, the latent motion rule beneath hurricane trajectories is figured out and a certain hurricane trend chart is printed by analyzing sub-trajectories of past hurricane trajectories in \cite{ferreira2013vector}. Sub-trajectories also lead to simplified trajectories which represent trajectory data as some smaller, less complex primitives suitable for storage and retrieval purposes \cite{anagnostopoulos2006global}. In \cite{yan2010hybrid}, sub-trajectories are generated by well-defined policies based on facility performance, time range or distance range. In \cite{bashir2007object} and \cite{bashir2007real}, trajectory is segmented at the so-called significant changing points at which direction or speed changes dramatically. Curvature describes direction information, and it could be extracted if a trajectory is treated as a curve by connecting consecutive trajectory points. Curvatures are computed by transforming 3-dimensional position coordinates of points into spherical system and quantized as {\emph{up, down, left, right}} \cite{faria20093d}, then a trajectory is segmented at the points where curvature changes. In addition, Minimum Description Length (MDL) principle traces the sub-trajectories holding primary movement of trajectory data by minimizing the differences between sub-trajectories and the corresponding trajectories in \cite{lee2007trajectory}. Minimum Bounding Rectangles (MBR) is proposed to separate trajectories under occlusion and optimize the inter-object separability in \cite{anagnostopoulos2006global}. It optimizes the bounding rectangles containing sub-trajectories to ensure that the distance between two rectangles are closer than the distance of trajectories.\\
\indent Some specific regions of surveillance area hold special semantic information and attract more attention so Regional Segmenting method is implemented. The whole scene is split into several regions and boundaries of the regions segment trajectories \cite{zhang2007trajectory}. As independent motion pattern, sub-trajectories characterize more information while original trajectory presents limited information.\\
\subsubsection{Points of Interest}
\indent Some specific regions of surveillance area hold special semantic information. Thus, the points inside the special regions are used to represent trajectory or scene in \cite{tuytelaars2008local} and all these points are called Points of Interest (POI). The points outside the regions are ignored because they are short of useful information. For instance, activity analysis is a key part in surveillance application to seek low-level situational awareness by understanding and characterizing behaviors of objects in the scene \cite{morris2008survey}, so it is critical to extract POI in the special regions. In topographical map, POI inside the special regions are represented as a single node. For example, two types of POI are introduced in \cite{morris2008survey} where the first one is the points in entry/exit zones and the second one is the points at the scene landmarks that objects intend to approach, move away or stay for a long time. Except for the special areas, points are represented by a node if their speed are less than a threshold in \cite{brandle2006track} and \cite{morris2011trajectory}. The importance of points can be measured and high-scored ones are selected in \cite{zhou2013dense}. For video data, POI are obtained by Pyramid Representation \cite{wang2011action}. In addition, optical flow is another popular implementation by estimating trajectory motion in \cite{wang2016robust} and \cite{fragkiadaki2015learning}.\\
\subsubsection{Scale-invariant Features}
\indent In image frames, more robust and representative features are needed rather than only positional information of trajectory points in \cite{jiang2012trajectory} and \cite{wang2011action}. In \cite{wang2011action}, histograms of oriented gradients (HOG) and histograms of optical flow (HOF) features are used to describe static appearance information and local motion information of trajectories, respectively. HOG feature computes orientation information to keep scale-invariant property of tracking point and it is fast to implement \cite{jiang2012trajectory}\cite{klaser2008spatio}\cite{laptev2008learning}\cite{willems2008efficient}\cite{matikainen2009trajectons}\cite{matikainen2010representing}. Furthermore, Scale-invariant Feature Transform (SIFT) descriptor represents image patch around tracking point \cite{uemura2008feature}\cite{sun2009hierarchical}\cite{wang2008video}\cite{vig2012space}\cite{wang2013dense}, and computes scale and orientation information of image patches to localize tracking object in consecutive frames. In \cite{uemura2008feature}, Kanade Lucas Tomasi (KLT) tracker is used to find trajectory points and SIFT is applied to represent them. In \cite{wang2008video}, Difference-of-Gaussian (DOG) detector is used to detecting trajectory points instead of KLT in \cite{uemura2008feature}.\\
\subsection{Common Distance Measurements}
\indent Essentially, trajectory are allocated into cohesive groups according to their mutual similarities. An appropriate metric is necessary \cite{morris2009learning}\cite{zhang2006comparison}\cite{atev2010clustering}.\\
\indent {\emph{Euclidean Distance}}: Euclidean distance requires that lengths of trajectories should be unified and the distances between the corresponding trajectories points should be summed up,\\
\begin{equation}
D(A,B)=\frac{1}{N}\sum[(a_n^x-b_n^x)^2+(a_n^y-b_n^y)^2]^{\frac{1}{2}},
\end{equation}
where $a_n^x$ and $a_n^y$ indicate the $n$th point of trajectory $A$ on Cartesian coordinate. $N$ is the total number of points. In \cite{nanni2006time}, Euclidean distance is used to measure {\emph{the contemporary instantiations}} of trajectories.\\
\indent {\emph{Hausdorff Distance}}: Hausdorff distance measures the similarities by considering how close every point of one trajectory to some points of the other one, and it measures trajectories $A$ and $B$ without unifying the lengths in \cite{liu2014entropy}\cite{chen2011clustering},\\
\begin{equation}
D(A,B)=\max\{d(A,B),d(B,A)\},
\end{equation}
\begin{equation}
\left\{
\begin{aligned}
d(A,B)&=\max_{a\in A}\min_{b\in B}||a-b||\\
d(B,A)&=\max_{b\in B}\min_{a\in A}||b-a||,
\end{aligned}
\right.
\end{equation}
\indent {\emph{Bhattacharyya Distance}}: Bhattacharyya distance measures how closely of two probability distributions. In \cite{li2006coarse}, it is employed to measures similarities of quantized directions of points,\\
\begin{equation}
D(A,B)=-\ln (BC(A,B)),
\end{equation}
where $BC(A,B)=\sum_{t=1}^T\sqrt{a_t\cdot b_t}$ and it is used to measure the separability of $A$ and $B$. $a_t$ and $b_t$ are quantized directions.\\
\indent {\emph{Frechet distance}}: Frechet distance measures similarity between two curves by taking into account location and time ordering. After obtaining the curve approximations of trajectories $A$ and $B$, their curves map unit interval into metric space $S$, and a re-parameterization is added to make sure $t$ cannot be backtracked. Frechet distance is defined as\\
\begin{equation}
D(A,B)=\inf_{\alpha,\beta}\max_{t\in [0,1]}\{d\left(A(\alpha(t)),B(\beta(t))\right)\},
\end{equation}
where $d$ is distance function of $S$, $\alpha$, $\beta$ are continuous and non-decreasing re-parameterization.\\
\indent {\emph{Dynamic Time Warping (DTW) Distance}}: DTW is a sequence alignment method to find an optimal matching between two trajectories and measure the similarity without considering lengths and time ordering \cite{shao2016integral}\cite{bautista2016gesture}.\\
\begin{equation}
W(A,B)=\min_f\frac{1}{n}\sum_{i=1}^n||a_i-b_{f(i)}||_2,
\end{equation}
where $A$ has $n$ points and $B$ has $m$ points, all mappings $f:[1,n]\to [1,m]$ should satisfy the requirements that $f(1)=1,\ f(n)=m$ and $f(i)\leq f(j),\ for\ all\ 1\leq i\leq j\leq n$.\\
\indent {\emph{Longest Common Subsequence (LCSS) Distance}}: LCSS aims at finding the longest common subsequence in all sequences, and the length of the longest subsequence could be the similarity between two arbitrary trajectories with different lengths. The distance $LCSS_{\epsilon,\delta}(A,B)$ is written as\\
\begin{equation}
LCSS_{\epsilon,\delta}(A,B)=\left\{
\begin{aligned}
&0,\ if\ A\ or\ B\ is\ empty\\
&1+LCSS_{\epsilon,\delta}(Head(A),Head(B)),\\
&\qquad if\ ||a_N-b_M||<\epsilon\ and\ |N-M|<\delta\\
&max(LCSS_{\epsilon,\delta}(Head(A),B),\\
&\qquad LCSS_{\epsilon,\delta}(A,Head(B))),\ otherwise,
\end{aligned}
\right.
\end{equation}
where $Head(A)$ indicates first $N-1$ points belonging to $A$ and $Head(B)$ denotes first $M-1$ points of $B$. Finally, $D(A,B)=1-\frac{LCSS_{\epsilon,\delta}(A,B)}{\max(N,M)}$.\\
\indent In \cite{lee2007trajectory}\cite{lee2008traclass}\cite{li2010incremental}, more other distance types are proposed to consider more properties such as angle distance, center distance and parallel distance, which are defined as\\
\begin{equation}
d_{angle}(L_i,L_j)=\left\{
\begin{aligned}
&||L_j||\times \sin (\theta),\ 0^o\leq \theta\leq 90^o\\
&||L_j||,\ 90^o\leq \theta\leq 180^o,
\end{aligned}
\right.
\end{equation}
where $\theta$ is the smaller intersecting angle between $L_i$ and $L_j$.\\
\begin{equation}
d_{center}(L_i,L_j)=||center_i-center_j||,
\end{equation}
where $d_{center}(L_i,L_j)$ is the Euclidean distance between center points of $L_i$ and $L_j$.\\
\begin{equation}
d_{parallel}(L_i,L_j)=\min(l_1,l_2),
\end{equation}
where $l_1$ is the Euclidean distances of $p_s$ to $s_i$ and $l_2$ is that of $p_e$ to $e_i$. $p_s$ and $p_e$ are the projection points of $s_j$ and $e_j$ onto $L_i$ respectively.\\
\indent Distance metrics are used in much more fields relating to trajectories clustering, e.g., density clustering \cite{lee2007trajectory}\cite{lee2008traclass}\cite{palma2008clustering}\cite{bu2009efficient}\cite{andrienko2009interactive}. It is critical to choose an optimal distance according to the scene. For instance, LCSS distance is proved to provide outperforming performance without concerning trajectories length \cite{morris2009learning}. Hausdorff distance aims at finding the minimum distance between two trajectories and ignore time-order in data. A comparison of distance is listed in Table I,\\
\begin{table}[!htbp]
\caption{Summary of common distance measurements}
\begin{center}
\begin{tabular}{ccc}
\toprule
Measurement types & Unifying lengths & Computational complexity\\
\midrule
Euclidean distance& Yes& $O(n)$\\
Hausdorff distance& No& $O(mn)$\\
Bhattacharyya distance& Yes& $O(n)$\\
Frechet distance& No& $O(mn)$\\
LCSS distance& No& $O(mn)$\\
DTW distance& No& $O(mn)$\\
other distance types& No& $O(1)$\\
\bottomrule
\end{tabular}
\end{center}
\end{table}
\section{Unsupervised Algorithms of Trajectory Clustering}
\indent Unsupervised algorithms infer a function to describe internal relationships between unlabeled data. Clustering is the method to draw this hidden structure, and some models relating to trajectory clustering are reviewed such as Densely Clustering models, Hierarchical Clustering models and Spectral Clustering models.\\
\subsection{Densely Clustering Models}
\begin{figure} [!h]
\centering
\includegraphics[width=2.8in,height=1.8in]{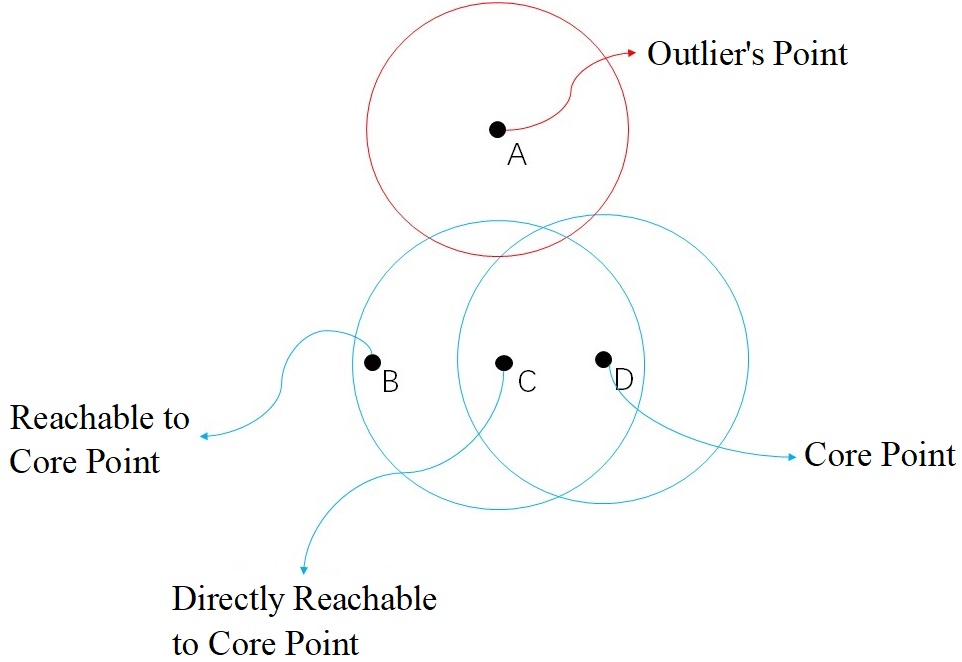}
\caption{DBSCAN}
\end{figure}
\begin{figure} [!h]
\centering
\includegraphics[width=3.8in,height=1.2in]{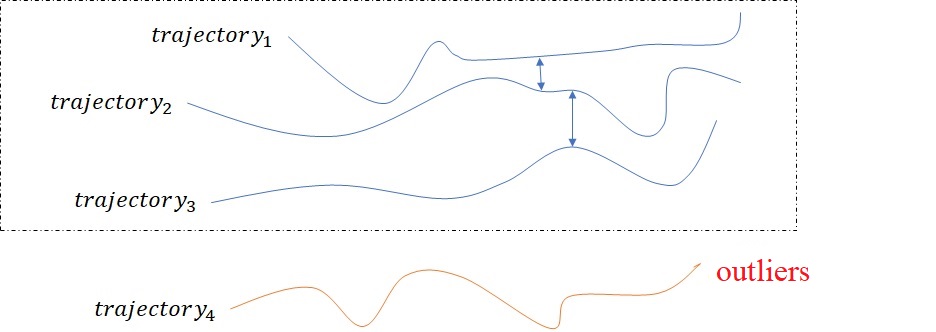}
\caption{DBSCAN for trajectory clustering}
\end{figure}
\indent Given the centroids, the closely points are packed together and this procedure is called densely clustering. Inspired by this idea, Density-based spatial clustering of applications with noise (DBSCAN) which has been widely applied to trajectory clustering is proposed in \cite{ester1996density}. In DBCSAN, point $p$ is chosen as the core point and distance threshold $\epsilon$ is given in advance. The points inside circle of which the radius is $\epsilon$ and the center is $p$ are called {\emph{directly reachable}} to $p$. Furthermore, points $\{q_1, q_2, ......, q_n\}$ are {\emph{reachable}} to $p$ if there is a path that $q_1$ is {\emph{directly reachable}} to $p$ and each $q_{i+1}$ is {\emph{directly reachable}} $q_i$ \cite{jeung2008discovery}\cite{wang2013action}. Other points are the outliers. Thus, the distance metric and the core parts selection are important. For solving the problem that DBSCAN cannot cluster the trajectories with large differences in densities \cite{lee2007trajectory}\cite{lee2008traclass}\cite{khaing2014efficient}, all trajectories are partitioned and substituted by sub-trajectories, then sub-trajectories are clustered and all clusters are grouped at the last step. However, different from measuring distance by Euclidean distance in \cite{lee2007trajectory}, the distance is measured by a combination of angle distance, center distance, parallel distance with equal weight in \cite{lee2008traclass} and \cite{khaing2014efficient}. The core trajectories are computed from the clusters and used for classifying new coming trajectory in \cite{lee2008traclass}, \cite{zheng2015trajectory}, \cite{deng2011trajectory} and \cite{zheng2010geolife}, e.g., all trajectories points belonging to same cluster are averaged as a new point at each time, and all averaged points form the representations of clusters \cite{lee2008traclass}. In an adaptive multi-kernel-based method, shrunk clusters represent all groups by considering the attributes including positions, speeds and points, which retains much more discriminative messages in \cite{xu2015unsupervised}.\\
\indent Besides DBSCAN, there are some other models belonging to Densely Clustering models cluster trajectory data. K-means clusters trajectories by searching centroids of clusters repeatedly \cite{ferreira2013vector}\cite{morris2011trajectory}\cite{suzuki2007learning}\cite{galluccio2012graph}\cite{hu2006system}\cite{melo2006detection}. For improving the performance of K-means, Expectation Maximization (EM) algorithm is implemented to solve optimization problem iteratively \cite{zhou2007detecting}. Due to the issues such as data imprecision and complexity of large data sets, a trajectory may belong to multiple clusters so EM is used to classify them \cite{kim2008real}. Fuzzy C-Means (FCM) algorithm employs parameters to measure the level of cluster fuzziness for each trajectory, called fuzzifier. The algorithm searches correct direction in each iteration for cluster trajectories \cite{pelekis2009clustering}\cite{pelekis2011clustering}\cite{simonnet2017multi}.\\
\subsection{Hierarchical Clustering Models}
\begin{figure} [!h]
\subfloat[Agglomerative clustering]{
\includegraphics[width=2.5in,height=1.4in]{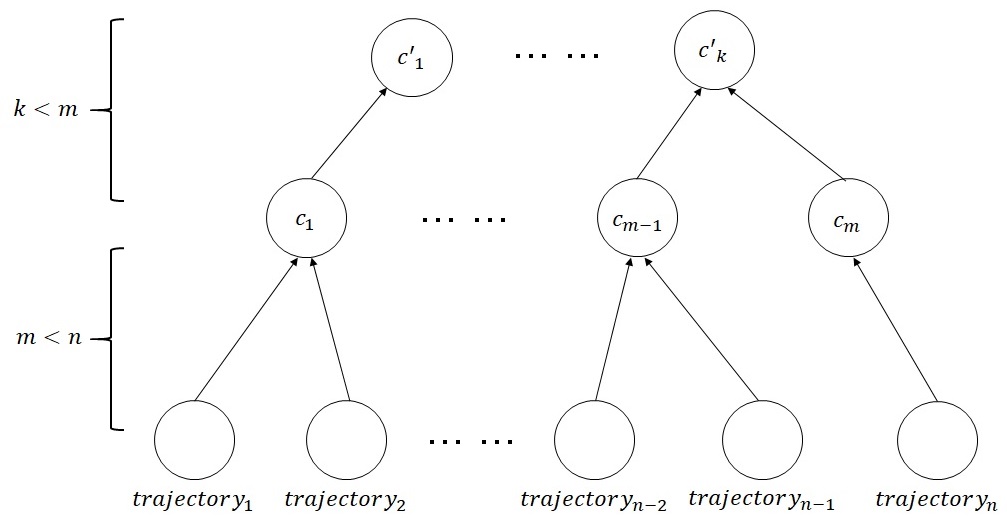}}
\subfloat[Divisive clustering]{
\includegraphics[width=2.5in,height=1.4in]{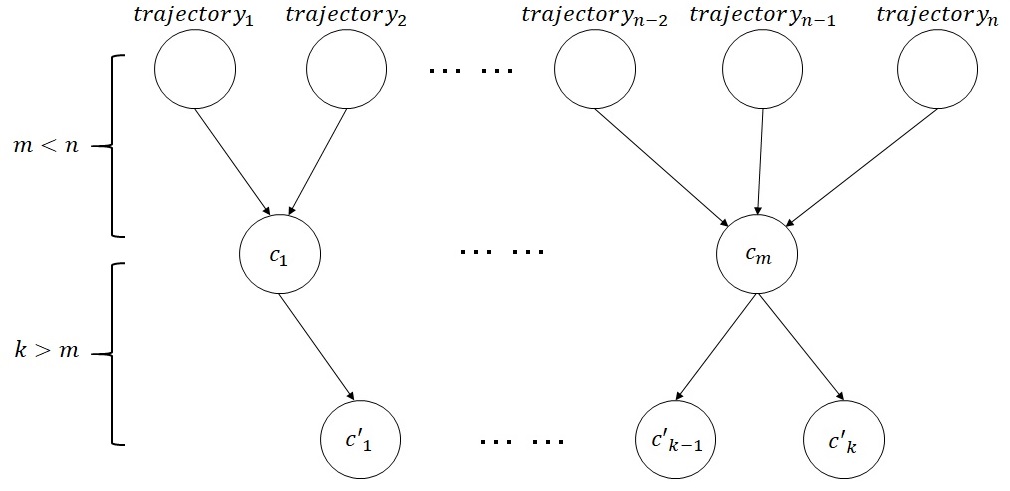}}
\caption{Hierarchical clustering models}
\label{fig6}
\end{figure}

\indent Hierarchical Clustering models help to understand trajectory by multiple features, so this tree-type construction is proper to implement. Hierarchical Clustering models generally fall into two clustering types, Agglomerative and Divisive. As shown in Fig.\ref{fig6}, two hierarchical types are also known as ``bottom-up'' and ``top-down'' approaches.\\
\indent In Agglomerative frameworks, trajectories are grouped and the similar clusters are merged by searching their common properties. Optimal classifications are obtained by repeating representation computation and clusters merging until meeting the requirements. Inspired by this idea, Agglomerative clustering models were explored in \cite{zheng2009mining} to mine the locations that users are interested, Hypertext Induced Topic Search (HITS) model is proposed to achieve this goal and movement tracks of users are recorded as trajectories. Top $n$ interesting trajectory clusters are obtained iteratively and the most popular locations are generated.\\
\indent Different from Agglomerative, Divisive frameworks cluster trajectory data into groups and split them recursively to reach the requirements. Following this framework, trajectory data are characterized by direction feature and clustered by Dominant-set embedded Bhattacharyya distance in initial clustering stage \cite{li2006coarse}. In each cluster, trajectories are split further except for the ones holding similar positions. Because of the good performance of iterative models, Test-and-Divide (TAD) model is proposed \cite{zheng2013discovery}. It is a Divisive framework detecting all the closed trajectories firstly and splitting them recursively. More attributes of trajectory points are considered to improve the performance in \cite{wang2006learning}. For instance, trajectory $A=\{a_1,a_2,\cdots,a_n\}$ where $a_i=\langle x_i,y_i,\beta_i\rangle$. It is comprised of 2-dimensional position and an additional attribute $\beta$ such as velocity or object size. In the coarse clustering step, the distance measurement between trajectory $A$ and its nearest observation trajectory $B$ are shown as follows,\\
\begin{equation}
f(A,B)=\frac{1}{N_A}\sum_{a_i\in A}||(x_i^a-x_{\psi(i)}^b,y_i^a-y_{\psi(i)}^b+\gamma d(\beta_i^a,\beta_{\psi(i)}^b))||,
\end{equation}
where $\psi(i)=\arg\min_{j\in B}||(x_i^a-x_j^b,y_i^a-y_j^b)||$ and the minimum distance value is counted as the distance between $A$ and $B$. $N_A$ is the total number of points belonging to $A$, $d(\beta_i^a,\beta_{\psi(i)}^b)$ indicates the dissimilarity of $A$ and $B$, and $\gamma$ is weight parameter. In the fine-clustering stage, the model aims at distinguishing distortions by considering directed similarity $S_{A\to B}$ and confidence $C_{A\to B}$
\begin{equation}
S_{A\to B}=\frac{\sum_{a_i\in A}c(a_i,b_{\psi(i)})s(a_i,b_{\psi(i)})}{\sum_{a_i\in A}c(a_i,b_{\psi(i)})},
\end{equation}
\begin{equation}
C_{A\to B}=\frac{\sum_{a_i\in A}c(a_i,b_{\psi(i)})^2}{\sum_{a_i\in A}c(a_i,b_{\psi(i)})},
\end{equation}
where $c(a_i,b_{\psi(i)})=\exp(\frac{-||(x_i^a-x_{\psi(i)}^b,y_i^a-y_{\psi(i)}^b)||}{\sigma_1})$ and $s(a_i,b_{\psi(i)})=\frac{\exp(-d(\beta_i^a,\beta_{\psi(i)}^b)}{\sigma_2}$.\\
\indent Furthermore, a similar hierarchical framework is explored to group videos by constructing the trajectories of video \cite{gaidon2014activity} as an unordered tree, and a kernel method recognizes videos by clustering the trees. In addition, Hierarchical Clustering models also recognize actions from video in \cite{raptis2012discovering} and \cite{tabatabaei2011ganc}. For two trajectories in video, $a=\{x_1^a,x_2^a,\cdots,x_{t_a}^a\}$ and $b=\{x_1^b,x_2^b,\cdots,x_{t_b}^b\}$, the distance is computed as follows,\\
\begin{equation}
d(a,b)=\max_{t\in[\tau_1,\tau_2]}d_{spatial}[t]\cdot \frac{1}{\tau_2-\tau_1}\sum_{t=\tau_1}^{\tau_2}d_{velocity}[t],
\end{equation}
where $d_{spatial}[t]$ is the positional distance at time stamp $t$, and $d_{velocity}[t]$ is the similarity measurement of velocity. An affinity matrix $w(a,b)=exp(-d(a,b))$ is calculated and trajectories are clustered by greedy agglomerative hierarchical models \cite{raptis2012discovering}\cite{tabatabaei2011ganc}. The clusters are overlapped because of similar parts, so every trajectory is weighted and optimized to classify in \cite{ni2015motion}. Since one motion object may generate several trajectories, it is critical to employ as much features as possible to ensure object recognition, and a multi-layer classifier is invented in \cite{li2006coarse}\cite{antonini2006counting}.\\
\subsection{Spectral Clustering Models}
\indent Trajectory data can be represented as a matrix called affinity matrix, and the relationships between them are extracted as the elements of matrix. The top $K$ eigenvectors form clusters with distinctive gaps between them which can be readily used to separate data into different groups \cite{xiang2008spectral}. In addition, affinity matrix characterizes videos \cite{turchini2015understanding} and represents the relationships. In \cite{hu2007semantic}, affinity matrix $A$ is constructed as follows,\\
\begin{equation}
A_{ij}=\exp[\frac{-\bar{d_{ij}}}{2\sigma^2}],
\end{equation}
where $\bar{d_{ij}}=\frac{1}{n}\sum_{k=1}^n||x_{i,k}-x_{j,k}||$, and $x_{i,k}$ indicates the $k$th point of trajectory $i$. Considering different lengths of trajectories, some novel models are explored to construct affinity matrix \cite{brunskill2007topological}\cite{brox2010object} and it is constructed as\\
\begin{equation}
A_{ij}=\left\{
\begin{aligned}
& e^{(-\frac{1}{\sigma_i\sigma_j}||v_i-v_j||^2)}, \ for\ i\neq j\\
& 0, \ otherwise,
\end{aligned}
\right.
\end{equation}
where $v_i$ and $v_j$ are points, $\sigma_i$ and $\sigma_j$ indicates scale invariance computed by the median of the $l$ nearest neighbors. In order to increase the separation of points belonging to different groups, SVD decomposition is used to construct the affinity matrix \cite{lauer2009spectral}. In addition, a novel distance method is explored to compute trajectories $P$ and $Q$ \cite{atev2006learning} so that spatial distinction can be considered. \\
\begin{equation}
s(P,Q)=e^{-\frac{1}{2}h_\alpha(P,Q)h_\alpha(Q,P)/(\sigma_p\sigma_Q)},
\end{equation}
\begin{equation}
h_{\alpha}(P,Q)=ord_{p\in P}^{\alpha}\left( \min_{q\in N(C(p))} d(p,q) \right),
\end{equation}
where $h_{\alpha}(P,Q)$ is the directed Hausdorff distance, $ord_{p\in P}^{\alpha}f(p)$ indicates the value of $f(p)$ and $N(C(p))$ denotes the subset of points which the ones matching to the point $p$ in trajectory $P$.\\
\indent For clustering high dimensional trajectory data by Spectral Clustering models, several novel methods are explored in \cite{zhang2016red}, \cite{hong2017high} and \cite{chen2009spectral}. For example, a new similarity metric captures causal relationships between time series in \cite{hong2017high} and a mixture of affinity subspaces is applied to approximate trajectory in \cite{chen2009spectral}. Trajectory data are represented by considering covariance features of trajectories in \cite{ergezer2016anomaly}, so it avoids considering different lengths of trajectory data. Spectral clustering works with multiple-instance learning frameworks to achieve human action recognition in \cite{yi2016human}.\\
\indent Spectral Clustering models are derived from Graph Theory in which an undirected graph represents the relationships and constructs a symmetric adjacency matrix presenting them \cite{brouwer2011spectra}. By constructing a graph, both explicit and implicit intentions inside trajectory data are mined \cite{chen2015action}. The graph is cut into sub-graphs to classify trajectories, and each sub-graph represents its own cluster \cite{zhang2009learning}\cite{lin2012integrating}. Hierarchical layers search sub-clusters in each cluster by treating trajectories points as graph nodes and this procedure is called {\emph{Hierarchical graph partitioning}} \cite{guo2010graph}. For considering more variables, a novel measurement function comprised of the entropy rate of a random walk on a graph is presented in \cite{liu2014entropy}. From the idea that an undirected graph can be represented as an adjacent matrix, a directed graph also can be involved \cite{li2007traffic}. Trajectory Binary Partition Tree (BPT) represents video in \cite{palou2013hierarchical} by representing trajectories as nodes so the edges indicate relationships between a pair of trajectories, and graph cut method groups trajectory data. Because of the robustness of composite feature descriptors, the descriptors including Speeded Up Robust Features (SURF) and Maximally Stable Extremal Regions (MSER) are employed in \cite{lin2012integrating}. An object creates several trajectories if different parts of the object are tracked, so a model is invented to describe trajectories by feature patches \cite{liu2009trajectory}. The edges are computed by geometric distance and appearance distance. Hausdorff distance is utilized to measure the similarities and set as weights of edges in \cite{junejo2008euclidean}. Since the great performance of PageRank, it is used to score the edges in \cite{cho2010authority}, too.\\
\subsection{Discussion}
\indent Densely Clustering models classify trajectories by distance metrics mostly, which may result in classifying trajectory data by spatial information. Hierarchical Clustering models fix this problem by considering more attributes in each level. However, this operation cost much more time in computation. Spectral Clustering models compute internal relationships by analyzing the affinity matrix, and it saves much more computational resource by processing all trajectory data together. But \cite{keuper2015motion} mentions that Spectral Clustering models have their own limitation that they are well defined only for the non-negative affinities between trajectories. Furthermore, that trajectory lengths are required to be unified is another issue of applying Spectral Clustering models.\\
\section{Supervised Algorithms of Trajectory Clustering}
\indent Supervised algorithms aims at learning a function which determines the labels of testing data after analyzing labeled training data. Therefore, supervised algorithms outperform on scene understanding and utilization of ground truth, and the algorithms could save much more computation resource. In some supervised algorithms, trajectory data are classified by unsupervised algorithms and the representations of clusters are obtained to classify new inquiry trajectories. For example, in Densely Clustering models, the representations can be computed from the grouped training trajectory data and new coming trajectories are clustered quickly in \cite{bashir2007object} and \cite{pelekis2009clustering}. Trajectory data are classified and organized in a tree-construction and new coming trajectories are clustered by searching the tree in \cite{gurung2014traffic} and \cite{pelekis2017temporal}.\\
\subsection{Nearest Neighbor Algorithms}
\begin{figure} [!h]
\centering
\includegraphics[width=2.8in,height=1.3in]{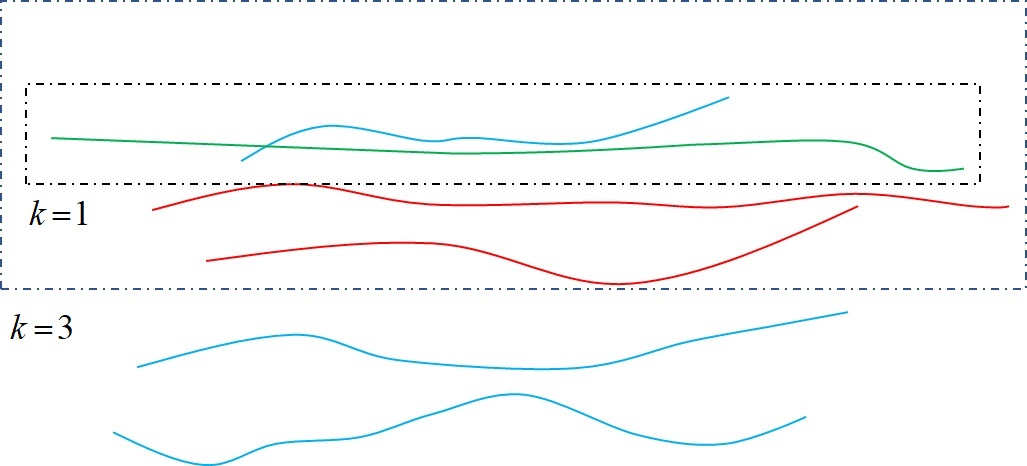}
\caption{$k$-NN for trajectory clustering. Inquiry trajectory is the green one, the labeled data are the red and the blue ones which means two clusters.}
\label{fig7}
\end{figure}

\indent Nearest Neighbor algorithms, such as $k$-Nearest Neighbor algorithm ($k$-NN), are finding a voting system to determine the category of a new coming entity and all data are kept in the same feature space. In trajectory clustering, the distances from an inquiry trajectory to all labeled trajectory data are computed, and the label of the inquiry trajectory is voted by its $k$ nearest neighbors. Shown in Fig.\ref{fig7}, the inquiry trajectory is assigned as blue cluster if $k=1$ and assigned as red one if $k=3$.\\
In the implementation, it is important to choose a suitable distance metric according to the scenario, occlusion, trajectory data sources and feature types. Therefore, trajectory data are represented by MBR and classified by $k$-NN in \cite{gao2007efficient}. It avoids occlusion and increasing inter-object separability. Furthermore, trajectory data are represented in Riemannian manifold \cite{devanne20153} so their shapes can be modeled and compared by using an elastic metric. For accessing $k$-NN faster, fast nearest neighbor (fastNN) algorithm organizes trajectory data in an Octree \cite{poularakis2016low}. With the increasing inquiry trajectories, the trends of trajectory data in a fixed period are required instead of general representation, so a circumstance that dynamically searches the nearest neighbors in a fixed period or the ones belonging to some specific types is considered in \cite{frentzos2007algorithms}. Support Vector Machine (SVM) is trained to generate the hypervolume, and the inquiry trajectory is determined as outliers if it falls outside the hypervolume \cite{piciarelli2008trajectory}. Structural Support Vector Machine (Structural SVM) is explored to detect social groups in crowds in \cite{solera2016socially}. Furthermore, SVM works with Graph Theory to cluster trajectories in \cite{singh2017graph}.\\
\subsection{Statistical Models}
\indent Statistical models exploit a set of probability distributions to represent the data generating process such as Gaussian Mixture model (GMM) and Bayesian inference. GMM usually combines with EM algorithm to train each component, and Bayesian inference obtains a set of probability functions which determine the categories of inquiry trajectory data. Bayes' theorem is critical for Bayesian inference and written as $P(A|B)=\frac{P(B|A)P(A)}{P(B)}$ where $A$ and $B$ indicate two events in event space.\\
\indent GMM aims at describing the sample from $\{x_1,x_2,\cdots,x_n\}$ in a component of GMM as\\
\begin{equation}
P(x_j)=\sum_{i=1}^K\pi_i N(x_j;\mu_i,\Sigma_i),
\end{equation}
where $N(x_j;\mu_i,\Sigma_i)$ is the probability density of the $i$th component belonging to a component with mean $\mu_i$ and variance $\Sigma_i$. $\pi_i$ is the weight with a constraint that $\sum_{i=1}^K\pi_i=1$, and they can be computed according to event frequency. Generally, EM algorithm iteratively optimizes the parameters of GMM, but Maximum Likelihood algorithm is implemented instead of EM if labeled trajectory data are available in training stage. For example, video events are treated as a linear combination of a set of event patterns, and two probabilistic terms are proposed to characterize video events in \cite{yuan2017statistical}. Furthermore, the abnormal patterns are scored by summarizing the probabilities of trajectory data of the corresponding video. GMM models the variance caused by the environmental factors and embedded into DTW to recognize gestures \cite{bautista2016gesture}.\\
\indent Bayesian inference classifies new coming data, and the classified ones update the probability functions of Bayesian inference. For samples $\{x_1,x_2,\cdots,x_n\}$, the probability of the corresponding labels $y_{1:n}$ is $p(y_{1:n}|x_{1:n})$. Derived from Markov Chain Monte Carlo (MCMC) algorithm, the distribution of variables can be approximated by a joint distribution, so Gibbs sampling is used to approximate $p(y_{1:n}|x_{1:n})$ by sampling $p(y_i|y_{-i}, x_{1:n})$ iteratively. According to Bayes' theorem, $p(y_i|y_{-i}, x_{1:n})$ is represented as $p(y_i|y_{-i}, x_{1:n})\propto p(x_i|y_i)p(y_i|y_{-i})$ where $p(x_i|y_i)$ is the likelihood and $p(y_i|y_{-i})$ is the marginal distribution. In Dirichlet Process (DP) model which is one of the Bayesian inference frameworks, $p(y_i|y_{-i})$ is formulated as $p(y_i|y_{-i})\propto \alpha G_0(y_i)+\sum_{j\in{-i}}\delta(y_i-y_j)$ where $\alpha$ is scale parameter and $G_0$ is base measure in sample space. The clusters can be parameterized for classifying new inquiry data, e.g., Dirichlet Process Mixture model (DPMM) is used to represent all $m$ clusters with parameterized indexes $\{\Theta_1, \Theta_2,\cdots,\Theta_m\}$ in \cite{hu2013incremental}. Finally, the new inquiry trajectory is classified by a trained DPMM as $p(\Theta_k|R)\propto p(R|\Theta_k)p(\Theta_k)$ where $p(R|\Theta_k)$ is the likelihood and $p(\Theta_k)$ is the prior probability. In order to learn coupled spatial and temporal patterns, Hierarchical Dirichlet Process (HDP) algorithm is applied in \cite{wang2016globally}. Bayesian model is used to segment object by classifying trajectories, so that human motion is also detected \cite{devanne2017motion}.\\
\subsection{Neural Network}
\indent Neural network is an artificial system simulating the biological neural network in animal brains. The network is constructed by a number of mutually connected neurons, and each neuron is represented as a real number. Neural networks can represent data such as deep generative model. It is trained to represent multivariate time series if trajectory data are generated as a vector \cite{yuan2016deep}, and a deep fully-connected Neural Network with weight decay and sparsity constraint transfers trajectory data from different viewpoints to a fixed viewpoint in compact representation \cite{rahmani2018learning}.\\
\begin{figure*} [!htb] 
\centering
\includegraphics[width=5in,height=1in]{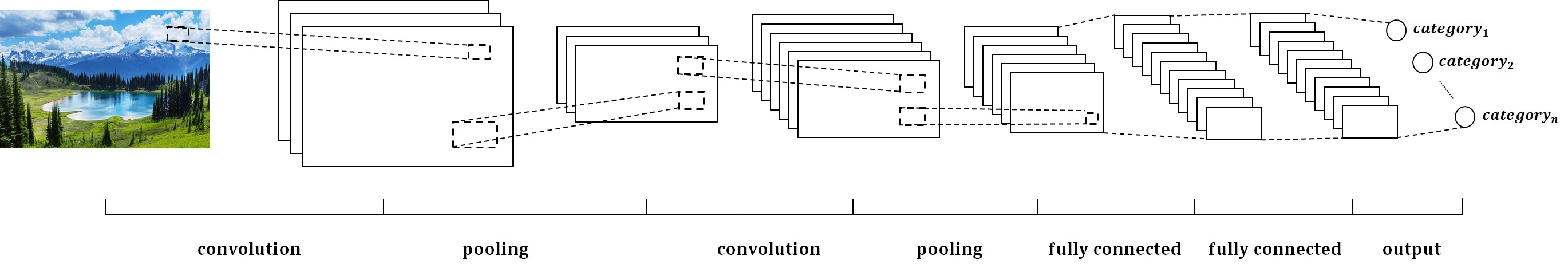}
\caption{CNN is one of the classical models of Neural Network and widely used in images classification}
\label{fig9}
\end{figure*}
\indent In most cases, Neural Network is used to classify data. It can be viewed as a mathematical function $f:X\to Y$ where $X$ is the observation and $Y$ indicates the corresponding label. For example, Convolutional Neural Network (CNN, or ConvNet) consists of multiple layers including convolutional, pooling and fully connected layers. That layout tolerates the variations of the input data, avoids overfitting problem and distinguishes data as similar as Multilayer perceptron (MLP). CNN has been proved efficient in clustering issue of computer vision. As Fig.\ref{fig9} shown, CNN is comprised by two convolutional layers, two pooling layers, two fully connected layers and one output layer which acts as an image classifier. CNN is employed for trajectory clustering in \cite{wu2016harnessing} and \cite{cho2014classifying}. Furthermore, CNN also ranks the trajectory clustering results in \cite{fragkiadaki2015learning}. A flexible deep CNN called Deep Event Network (DevNet) is trained by ImageNet data set, and the trained DevNet is tuned to extract generic image-level features of trajectory data in \cite{gan2015devnet}. In order to figure out the differences between image classification and multimedia event detection, DevNet fine tunes parameters by a specific data set, and backward passing is employed to identify pixels in consecutive frames to recount events. Deep Neural Network (DNN) is another Neural Network which learns a more compact and powerful representation of trajectories \cite{hasan2015continuous}. Furthermore, DNN keeps the structural relationships between trajectories in \cite{shi2015learning}, and mines the relationship between multiple features including spatiotemporal features, audio features and inter-class relationship to classify videos in \cite{wu2014exploring} and \cite{jiang2015exploiting}. Self-Organizing Map learns the similarities between trajectories in a 2-dimensional grid and each element of the grid indicates a specific prototype in \cite{naftel2006motion} and \cite{schreck2009visual}. In training steps, each training trajectory is trying to find the most suitable prototype in network, and adjust the neighbors of the matched one accordingly.\\
\subsection{Discussion}
\indent Nearest Neighbor algorithm only considers the spatial relationships between a pair of trajectory data but ignores local characters. Statistical model makes up for this imperfection by combining them in a mixture model or inferring the relationships in Bayesian models. Neural Network considers the differences of trajectory data and requires a huge number of data to train it. Though the supervised methods obtain the classifiers by observing a number of training data, overfitting problem may happen when the model overreacts training data.\\
\section{Semi-supervised Algorithms of Trajectory Clustering}
\indent Semi-supervised algorithms fall between unsupervised algorithms and supervised algorithms. The algorithms make use of a small number of labeled data and continuous inquiry data to complete tasks. The model is trained by labeled data firstly, then inquiry data are kept sending to the trained model to make sure that it can be updated to outperform the previous model. Semi-supervised procedure needs only a small cost in terms of human classification efforts. This procedure not only avoids overfitting problem, but also is more accurate than the unsupervised ones.\\
\indent Therefore, some semi-supervised algorithms are invented from unsupervised or supervised algorithms. For example, trajectory data are classified firstly and the new inquiry ones are clustered to update the classifier automatically \cite{gurung2014traffic}\cite{wang2014detection}\cite{yuan2017statistical}\cite{laxhammar2014online}. Detected anomaly trajectory data are used to recalculate the representation of anomaly trajectory clusters in \cite{laxhammar2014online}. Trajectory data of video are modeled as the combination of normal and abnormal patterns, and probabilistic terms characterize the patterns in \cite{yuan2017statistical}. From this modeling, the terms can be updated by the detected inquiry trajectory. In order to detect abnormal trajectories faster in complex scene, low-rank approximation is employed to describe trajectory data and the new detected abnormal ones update the threshold in \cite{wang2014detection}.\\
\indent Inspired by Hierarchical Frameworks, trajectories and the clusters are represented as a tree where children nodes indicate trajectories and roots denote the representations of the clusters in \cite{piciarelli2006line}, \cite{kulic2008incremental} and \cite{li2013effective}. A new cluster is created if no clusters close to the inquiry trajectory. Trajectory $T$ is constructed as a vector of 2 dimensional coordinates $T=\{t_{1}, \cdots, t_{n}\}$ where $t_{j}=\{x_{j}, y_{j}\}$. A representation of cluster is computed as $C_i=\{c_{i1}, \cdots, c_{im}\}$, where $c_{ij}=\{x_{ij}, y_{ij}, \sigma^2_{ij}\}$ and $ \sigma^2_{ij}$ is an approximation of the local variance of the cluster $i$ at time $j$. The inquiry trajectory is assigned to the nearest cluster and the corresponding cluster should be updated by the new one. For the nearest cluster point $c=\{x,y,\sigma^2\}$ to the point of trajectory $t=\{\hat{x},\hat{y}\}$, $c$ is updated as following\\
\begin{equation}
\left\{
\begin{aligned}
&x=(1-\alpha)x+\alpha\hat{x}\\
&y=(1-\alpha)y+\alpha\hat{y}\\
&\sigma^2=(1-\alpha)\sigma^2+\alpha[dist(t_i,c_j)]^2,
\end{aligned}
\right.
\end{equation}
where $\alpha$ is the update rate between 0 and 1.\\
\indent Considering the fact that Bayesian model is derived from Bayes' theorem, the parameters are optimized by sampling training data, and it is feasible to update the model by classified new inquiry data \cite{hu2013incremental}. Furthermore, in order to add new trajectory data, the previous samples and the new ones are sampled by Gibbs Sampling as\\
\begin{equation}
p(\eta_i|\eta_{-i},y_{1:N+\phi}) = p(\eta_i|\eta_{1:N}=W_{1:N},\eta^{new}_{-i},y_{1:N+\phi}),
\end{equation}
where $y$ is trajectory data, $\eta_{1:N}$ indicate the known states of the previous samples, and $N+1<i<N+\phi$. $\eta^{new}_{-i}$ denote the states of new inquiry trajectory data except for the $i$th one. From Bayes' theorem, the cluster process is rewritten as $p(y_i|\eta_i)p(\eta_i|\eta_{1:N}=W_{1:N},\eta^{new}_{-i})$. $p(y_i|\eta_i)$ is estimated by the previous samples and it is assumed to be Gaussian distribution. The only issue need to be fixed is carrying out Gibbs Sampling on $\eta_{N+1:N+\phi}$ to compute $p(\eta_i|\eta_{1:N}=W_{1:N},\eta^{new}_{-i})$.\\
\section{Promising Future Directions and Tasks}
\indent In trajectory clustering, we saw that how to measure trajectories with different lengths is important. Thus, a representation method or feature descriptor is essential for trajectory clustering. In recent years, transforming trajectory data into other space are paid more attention, such as DFT which keeping data information and unifying lengths of trajectory data \cite{hu2013incremental}. For other preparation works, re-sampling is efficient for sparse scene \cite{wang2009evaluation}, but it limits the robustness of model. Curve approximation fits the movement of trajectory \cite{zhang2009learning}\cite{sillito2008semi}. Hence, trajectory data preparation may be a promising and helpful direction.\\
\indent Recently, Densely Clustering models have achieved great progress in trajectory clustering. In particular, novel distance metrics have been proposed to measure trajectory data according to different properties. Furthermore, for the trajectory data with large difference in density, grid construction is employed to improve the performance \cite{uncu2006gridbscan}. Besides grid-based DBCSAN, sub-trajectories are acted as the substitute for trajectory in \cite{lee2007trajectory}, \cite{lee2008traclass} and \cite{khaing2014efficient}.\\
\indent Though Spectral Clustering models and Graph method share a similar idea, they are intrinsically different. Spectral Clustering models are easy to implement and have no restriction on data dimensions, but the models require non-negative affinities and this limitation restricts the performance and the application. Therefore, a suitable affinity matrix construction method is needed. Furthermore, it is critical to determine scale value when the affinity matrix is being computed, because it determines the clustering is fail or not. Thus, Spectral Clustering models need to handle the problem of constructing affinity matrix.\\
\indent In supervised algorithms, a large number of training data are required to obtain an efficient model. However, such as in Neural Network, there may have overfitting problem and some special steps are needed like pooling layers in CNN. In addition, it should be noticed that a meaningful distance metric is essential for Nearest Neighbor algorithms.\\
\section{Conclusion}
\indent In this paper, we reviewed the methods of trajectory clustering. They are classified into three categories: unsupervised, supervised and semi-supervised algorithms. Unsupervised algorithms can be grouped into three sub-categories: Densely Clustering models, Hierarchical Clustering models and Spectral Clustering models. Supervised algorithms are divided into Nearest Neighbor algorithms, statistical models and Neural Network. By means of a comprehensive analysis, we found that unsupervised algorithms have the disadvantages of high computation cost and heavy memory load, although there is no training data requirement and human experts supervising. Semi-supervised algorithms combine the advantages of both previous algorithms, and may result in more efficient methods. Finally, we proposed several promising future directions and tasks, and this paper could help readers to gain a thorough understanding of trajectory clustering.\\
\section*{References}
\bibliography{trajectory_view}
\end{document}